\documentclass[11pt,a4paper]{article}
\usepackage[hyperref]{naaclhlt2019}
\usepackage{times}
\usepackage{latexsym}

\usepackage[utf8]{inputenc} 
\usepackage[T1]{fontenc}    
\usepackage{hyperref}       
\usepackage{url}            
\usepackage{booktabs}       
\usepackage{amsfonts}       
\usepackage{nicefrac}       
\usepackage{microtype}      

\usepackage{graphicx}
\usepackage{subfigure}

\usepackage{amsmath}
\usepackage{amssymb}

\usepackage{multirow}

\usepackage{color}
\usepackage{xcolor}

\usepackage[ruled]{algorithm2e}

\usepackage{listings}
\usepackage{fancyvrb}

\DeclareMathOperator*{\argmin}{arg\,min}

\newcommand{\base}{\textsf{Base}}
\newcommand{\bv}{\textsf{Base-V}}
\newcommand{\adt}{\textsf{ADT}}
\newcommand{\adtd}{\textsf{ADT\textsubscript{D}}}
\newcommand{\adti}{\textsf{ADT\textsubscript{I}}}
\newcommand{\adtr}{\textsf{ADT\textsubscript{R}}}
\newcommand{\adts}{\textsf{ADT\textsubscript{S}}}
\newcommand{\adta}{\textsf{ADT\textsubscript{A}}}
\newcommand{\adtv}{\textsf{ADT}-V}
\newcommand{\adtdv}{\textsf{ADT\textsubscript{D}-V}}
\newcommand{\adtiv}{\textsf{ADT\textsubscript{I}-V}}
\newcommand{\adtrv}{\textsf{ADT\textsubscript{R}-V}}
\newcommand{\adtsv}{\textsf{ADT\textsubscript{S}-V}}
\newcommand{\adtav}{\textsf{ADT\textsubscript{A}-V}}

\newcommand{\ve}{\emph{VE}}

\aclfinalcopy 

\title{Word Shape Matters: \\Robust Machine Translation with Visual Embedding}

\author{Haohan Wang \and Peiyan Zhang \and Eric P. Xing\\
  School of Computer Science \\
  Carnegie Mellon University \\
  Pittsburgh, PA, USA \\
  {\tt haohanw@cs.cmu.edu} \\
  }

\date{}

\begin{document}
\maketitle

\begin{abstract}
Neural machine translation has achieved remarkable empirical 
performance over standard benchmark datasets, 
yet recent evidence suggests that 
the models can still fail easily dealing with  
substandard inputs such as misspelled words, 
To overcome this issue, 
we introduce a new encoding heuristic of the input symbols
for character-level NLP models:
it encodes the shape of each character 
through the images depicting the letters when printed. 
We name this new strategy \emph{visual embedding}
and it is expected to improve the robustness
of NLP models 
because 
human also process the corpus visually through printed letters, 
instead of machinery one-hot vectors. 
Empirically, 
our method
improves models' robustness 
against substandard inputs, 
even in the test scenario
where the models are tested with 
the noises that 
are beyond what is available during the training phase. 
\end{abstract}

\section{Introduction}
Despite the remarkable empirical successes 
Neural Machine Translation (NMT) has achieved 
over various benchmark datasets, 
models tested in the real-world scenario 
soon alarmed the community about 
the lack of robustness 
behind the appealing high numbers. 
For example, 
evidence shows that 
the impressive performances 
can hardly be met when the models 
are tested with out-of-domain or
noisy data \cite{luong2015stanford,belinkov2017synthetic,anastasopoulos2019neural},
while these datasets can barely 
raise any difficulties for human translators. 

This robustness disparity between human 
and NMT has motivated multiple recent works 
aiming to improve the 
resilience of NMT system against 
either natural or synthetic noises \cite{belinkov2017synthetic,zhou2019improving,vaibhav2019improving,sano2019effective,levy2019training}.
For example, 
most preceding works 
emphasized the importance of 
training with noisy inputs (\textit{i.e.}, adversarial training) 
\cite{belinkov2017synthetic,vaibhav2019improving,levy2019training}, 
which is further extended by \cite{sano2019effective} 
to perturb the intermediate representations other than input data. 

Different from the previous efforts along this line, 
we notice that 
previous NMT mostly process the corpus
through artificial one-hot vectors of words or characters, 
while human read through eyes, analyzing
the shape of the printed letters. 
We conjecture this disparity is the root 
of the notorious vulnerability of many NMT
towards substandard words,
such as misspellings and 
elongated words, 
which human can read through 
effortlessly. 

Therefore, 
in this paper, 
building around the central argument
{\begin{center}
\emph{Human read through eyes; so shall the models}\footnote{assuming the models are expected to show human-level resilience towards substandard inputs.}
\end{center}
}
we introduce a simple alternative of the one-hot character encoding mechanism
for character-level model:
we encode the shape of the symbols.  
This encoding of the letter shape, which we refer to as 
\emph{visual embedding} (\ve{}),
is obtained as a 
dimension-reduced representation 
of the images depicting the characters. 
The embedding exhibits 
remarkable robustness behavior against 
input noises in comparison to the previous standard. 
Our embedding also comes with a 
minor advantage as the embedding is 
even smaller than the typical one-hot 
embedding of characters, 
thus help to reduce the input size of the models.

The remainder of this paper is organized as follows. 
We first discuss the related works in Section~\ref{sec:related}, 
and then start to introduce our \emph{visual embedding} (\ve{}) and corresponding algorithms in Section~\ref{sec:method}. 
We demonstrate the empirical strength of methods in Section~\ref{sec:exp}, 
where we first verify the concept with synthetic data, 
and then compare against previous popular methods . 
We finally offer some discussion in Section~\ref{sec:diss} 
before we conclude our paper in Section~\ref{sec:con}. 

\section{Related Work}
\label{sec:related}
The vulnerability of modern neural networks 
towards human imperceptible input variations
has been studied for a while since \cite{szegedy2013intriguing}, 
primarily in the computer vision community (\textit{e.g.}, \cite{goodfellow2015explaining}), 
later extended to the NLP community (\textit{e.g.},\cite{ebrahimi2017hotflip,liang2017deep,yin2020robustness,jones2020robust,JiaRGL19,HuangSWDYGDK19,LiuMHXH19,PruthiDL19}).
Recent studies suggest that 
the fragility of neural networks 
roots in that 
the data has multiple signals that can reduce the 
empirical risk, 
and when a model is forced to reduce the training error, 
it picks up whatever information that
diminish the empirical loss, 
ignoring whether the learnt knowledge aligns with human 
perception or not \cite{wang2019high}, 
connecting the adversarial robustness problems 
and the bias in data problems 
that has been studied for a while 
(\textit{e.g.},\cite{wang2016select,goyal2017making,kaushik2018much,wang2019if}). 

NMT, 
despite the high empirical 
numbers on various leaderboards, 
complicated architecture design, 
and even occasional human-parity claims \cite{wu2016google,hassan2018achieving}, 
is fundamentally a statistical commission 
where a model is asked 
to search the patterns that reduce the 
empirical loss in training data 
and evaluated numerically in testing data, 
thus 
one may expect the models to ``cheat'' through 
distribution-specific signals for high scores and 
behave differently from human when tested more thoroughly \cite{laubli2018has}. 

As expected, 
Belinkov \textit{et al.}
showed that
some noises, either synthetic or natural, 
can easily break NMT models \cite{belinkov2017synthetic}, 
while human can process the noised texts effortlessly. 
Fortunately, after revealing 
this pitfall, 
Belinkov \textit{et al.} also proposed a simple
heuristic as a remedy \cite{belinkov2017synthetic}: 
augmenting the training samples with noises can 
significantly improve the robustness of these models against noises at test time. 
They also showed that  
the noises injected at training time and testing time have to be aligned to maximize the performance of this augmentation method. 

This augmentation method, 
also known as a member of the adversarial training family, 
has become the main force of the battle against the 
vulnerability issue of the NMT models \cite{levy2019training,sano2019effective,cheng2019robust}. 
For example, 
Levy \textit{et al.}
improved the robustness 
against character-level variations, such as typo, of the source languages \cite{levy2019training}.
They experimented with a transformer-based MT model with CNN character encoders, 
set up the experiment in a way that 
natural noises are only injected at testing time (so the evaluation result is a more natural reflection of what is in the real world), 
and observed that injecting synthetic noises during training help to improve the robustness against natural noises at testing. 

There are also other emerging forces 
against the fragility of NMT models. 
For example, 
inspired by a popular psychological observation suggesting that human are invariant to jumbled letters\footnote{
Most readers should be able to process the following text despite nearly all the words are jumbled: 
``Aoccdrnig to a rscheearch at Cmabrigde Uinervtisy, it deosn’t mttaer in waht oredr the ltteers
in a wrod are, the olny iprmoetnt tihng is taht the frist and lsat ltteer be at the rghit pclae.''
For more details, 
we refer readers to the online article linked at the next footnote.
}, 
Belinkov \textit{et al.} attempted the usage of 
average embedding of the characters compositing the word \cite{belinkov2017synthetic} 
and 
Sakaguch \textit{et al.}
introduced an architecture that predicts the original word back from the jumbled words \cite{sakaguchi2017robsut}. 
Recently, along another direction, 
Zhou \textit{et al.} demonstrated the resilence 
of a multitask learning mechanism against input variations \cite{zhou2019improving}. 

\emph{Key Differences:}
Our work is also inspired by the psychological observation discussed above.
However, instead of explicitly regularizing the models 
to fix what has been revealed, 
we ask the question that what can be 
a fundamental disparity between a machine and a human. 
Our answer, and the main argument of this paper, is that \emph{human read through eyes, thus the shape of the word matters more than the
exact permutation of the compositing letters.}
Interestingly, we notice that our argument 
is partially supported by an online post written by 
a psycholinguist commenting on the aforementioned observation\footnote{
\href{http://www.mrc-cbu.cam.ac.uk/people/matt-davis/cmabridge/}{www.mrc-cbu.cam.ac.uk/people/matt-davis/cmabridge/}}. 
Therefore, 
we design a new input regime that can 
encode the shape of the letters, 
and expect the shape 
of the words will be more accurately described 
by our technique. 

\section{Method}
\label{sec:method}
We first introduce the procedure that
can generate the \emph{visual embedding} (\ve{}) that 
describes the shape of the letters. 
The \ve{} is expected to work with
any existing 
character-level NLP models with no extra efforts, 
so we will only mention 
the corresponding neural architecture briefly.  

\subsection{Visual Embedding}
With a predetermined set of characters $\mathcal{C}$, 
we first choose a font (\textit{e.g.}, Times New Roman)
and decide the dimension of 
the image depicting a letter (\textit{e.g.} $15\times 15$ for each letter), 
then for every character in $\mathcal{C}$, we print it onto an image, which naturally offers us an embedding of the character (\textit{e.g.}, a vector of $225$, following our previous configuration). 
Further, to have a more memory-efficiency 
representation, 
we collect the embedding of all the characters in $\mathcal{C}$ and get a matrix of representations (\textit{e.g.}, a matrix of $|\mathcal{C}| \times 225$, following the previous configuration, where we use $|\cdot|$ to denote the cardinality of the set) 
and use PCA to map the matrix to a lower dimension $d$, which can be straightforwardly determined by inspecting the variance explained (details to follow). 
Therefore, we obtain representation as a $d$-dimension vector, which we name as the \ve{} of the character. 
Algorithm~\ref{alg:main} offers more formal details of this procedure. 

\paragraph{Determining the output dimension $d$ by examining ratio of explained variances.}
As the effective dimension of \ve{} is much smaller than the dimension of the image depicting the letters,
we use PCA to reduce the dimension to a smaller dimension $d$. 

We use a simple heuristic to decide an appropriate $d$: 
we plot the ratio of the explained variance of the first $d$ components calculated by PCA for every $d$, 
then we manually inspect the plot to select a cut-off
where the first $d$ components 
can sufficiently explain a large ratio (\textit{e.g.}, $95\%$) of the variance. 

The ratio of explained variance 
can be conveniently calculated by the ratio of the sum of the first $d$ eigenvalues over the sum of all the eigenvalues, where eigenvalues are calculated with the corresponding covariance matrix, thus guaranteed to be non-negative.

There are also other algorithms 
that can choose an appropriate $d$ automatically. 
However, 
since PCA of this scale of the data 
can be calculated in negligible time in a modern computer, and it only needs to be calculated once, 
we consider the plot-and-check procedure 
as the main method
to select an appropriate $d$. 
Also, the manually check procedure may be more reliable 
than automatic algorithms. 




\begin{algorithm}[t!]
\SetAlgoLined
\textbf{Input:} set of characters $\mathcal{C}$, font $T$, 
image dimension $m \times n$, output dimension $d$\;

\textbf{Output:} embedding of the shape $|\mathcal{C}| \times d$ (in other words, a length-$d$ vector for each character in the set)\;

Initialize a matrix $\mathbf{R}$ with shape $|\mathcal{C}| \times mn$\;

\For{every element $c$ in $\mathcal{C}$}{
    Initialize a blank image $\mathbf{I}_c$ of the size $m \times n$\;
    
    Print $c$ with font $T$ on $\mathbf{I}_c$\;
    
    Reshape the image into a vector $\mathbf{v}_c$ of dimension $mn$\;
    
    Collect $\mathbf{v}_c$ into the corresponding row of $\mathbf{R}$\;
}
Perform PCA to project $\mathbf{R}$ into a lower dimension matrix of the shape $|\mathcal{C}| \times d$
\caption{Algorithm of generating \emph{visual embedding}}
\label{alg:main}
\end{algorithm}

\subsection{Training with Visual Embedding}
\label{sec:model}

In simple words, to use \ve{}, all one need is to replace the traditional one-hot character input with \ve{}. 

\paragraph{Formal Discussion}
The data is a pair of collections of $n$ sentences, 
denoted as $\{\mathbf{X}, \mathbf{Y}\}$, 
where $\mathbf{X}$ denotes the source sentences 
and $\mathbf{Y}$ denotes the target sentences. 
The $i$\textsuperscript{th} sample of $\mathbf{X}$ (or $\mathbf{Y}$) is 
denoted as $\mathbf{x}_i$ (or $\mathbf{y}_i$) which 
consists a sequence of 
characters, 
denoted as 
$\{x_i^1, x_i^2, \dots, x_i^{l(i)}\}$ (or $\{y_i^1, y_i^2, \dots, y_i^{k(i)}\}$). 
$l(i)$ (or $k(i)$) denotes the number of characters
of the $i$\textsuperscript{th} source (or target) sentence. 
Sentences do not necessarily have the same length. 

Through the \ve{} technique, 
we will map the source sentence $\mathbf{X}$ (or $\mathbf{x}$, $x$)
into the representation $\mathbf{Z}$ (or $\mathbf{z}$, $z$, respectively). 
We use $e(\cdot;\theta)$ to denote the encoder and $\theta$ denotes 
its parameters;
we use $d(\cdot;\phi)$ to denote the decoder and $\phi$ denotes 
its paramters. 
The training of our NMT model is to optimize 
\begin{align*}
    \theta, \phi =\argmin_{\theta, \phi} \mathbb{E}_{\mathbf{x},\mathbf{y}} l(d(e(v(\mathbf{z}_i), \theta);\phi);\mathbf{y}_i)
\end{align*}
where $l(\cdot, \cdot)$ is a generic loss function and $v(\cdot)$ 
stands for the function mapping characters into \ve{}. 

\paragraph{Model Specification:} As our \ve{} mainly 
concerns with the character-level input, 
we discuss one corresponding 
network architecture and associated hyperparamters. 
Despite that the vectors can be integrated into 
almost any character-level models, 
the model used in this paper essentially builds upon 
transformer-based machine translation model \cite{vaswani2017attention}
and a CNN-based character encoder \cite{kim2016character} serving as the encoder, based on Fairseq implementation \cite{ott2019fairseq}.








\section{Experiments}
\label{sec:exp}
We first use synthetic experiment to validate the effectiveness of \ve{} for simple robust text classification, which is also a territory where we can discuss related questions such as the choice of a font and the choices of the effective dimension $d$ of \ve{}. 
Then we demonstrate the empirical strength of our method
with standard MT benchmarks when the test sentences are perturbed with noises. 

\subsection{Synthetic Experiment}

We first prove the concept of \ve{} with a simple binary text classification when the test sequences has some noises that are not seen by models during training. 

\paragraph{Experiment Setup} 
We generate the data by first sampling some positive ``words'' and negative ``words'', 
where each word is a sequence of three random letters. 
Each sample is a ``sentence'' of three ``words'', and the label (negative or positive) is determined by the majority of the polarity of the ``words''. 
Following this rule, we sampled 15 ``words'' for each category and generated roughly 180k samples for training and 60k samples for validation.
Further, 
in addition to the samples from the same distribution, 
we also generate two distributions of noised test samples: 
one is to mix random characters into the samples generated with the above rule, 
the other is to replace a ``word'' with a random sequence of three letters when the label of the ``sentence'' can still determined by the remaining two ``words''. 
Altogether, there are 60k samples for testing. 

The model we considered is inspired from the CNN-LSTM architecture \cite{vosoughi2016tweet2vec}, 
and we also implement the highway network \cite{srivastava2015training}. 
With this model, we compare the performance when the input is one-hot vector and is \ve{}. 

\paragraph{Results} 
The results are shown in Table~\ref{tab:synthetic}, where we report the performance of one-hot embedding and the results of \ve{} over four different fonts over nine different choices of the effective dimension $d$. 
First of all, we can observe the non-robustness issues of conventional one-hot embedding as the test accuracy is significantly lower than train accuracy. 
However, these issues are not observed for \ve{}. 
Further, we can see that the choices of fonts barely matter in the end performance, 
we believe this is mainly because all the fonts, 
despite the visual differences, 
are still essentially encoding the shape of the characters. 
Similarly, we can observe that the choices of the effective dimension $d$ barely matters for the end performance. 
With these evidence, the following experiments in this paper will use Time New Roman as the font and set $d=128$. 

\begin{figure*}
    \centering
    \subfigure[Arial]{\includegraphics[width=0.24\textwidth]{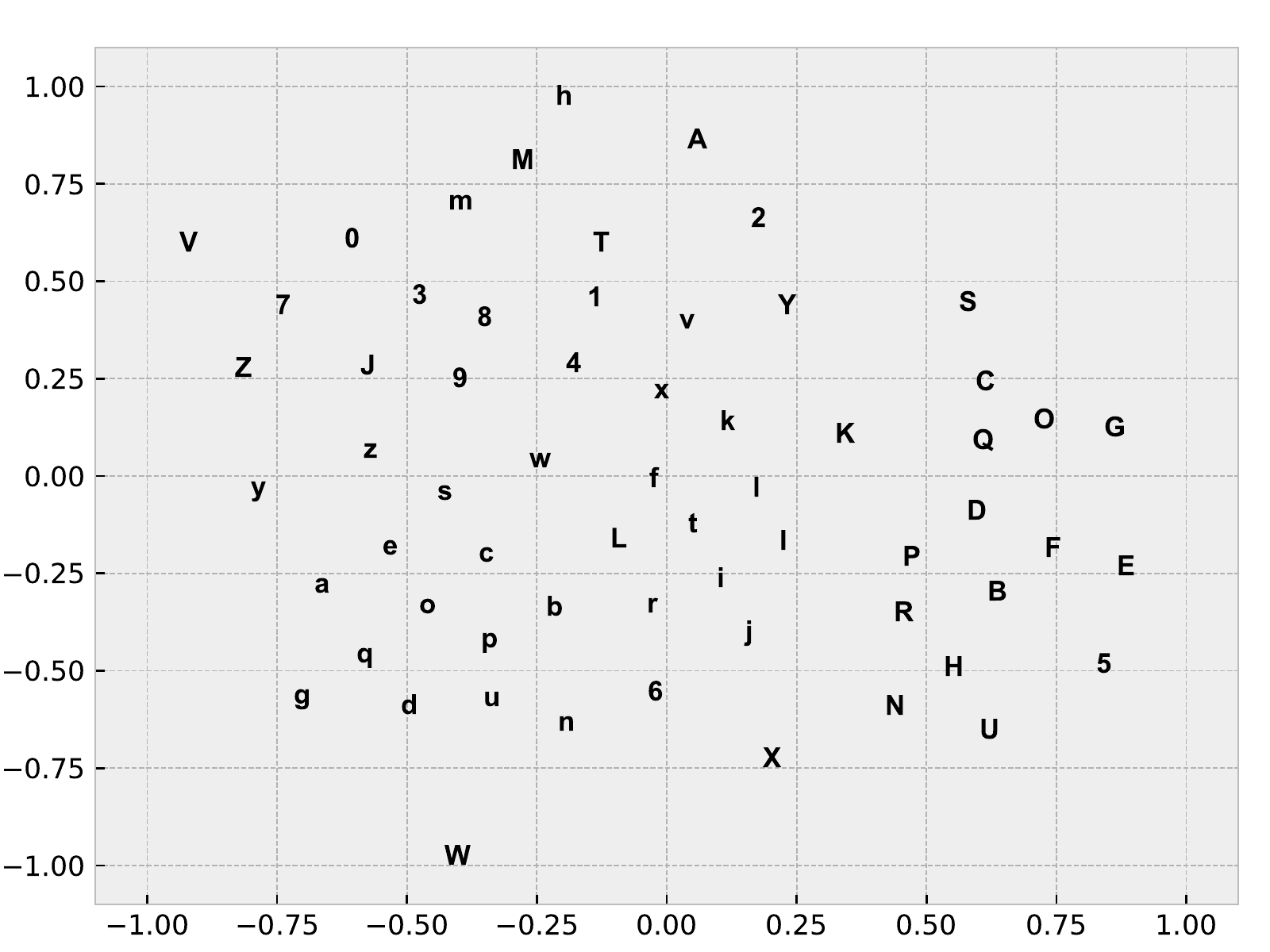}} 
    \subfigure[Comic Sans MS]{\includegraphics[width=0.24\textwidth]{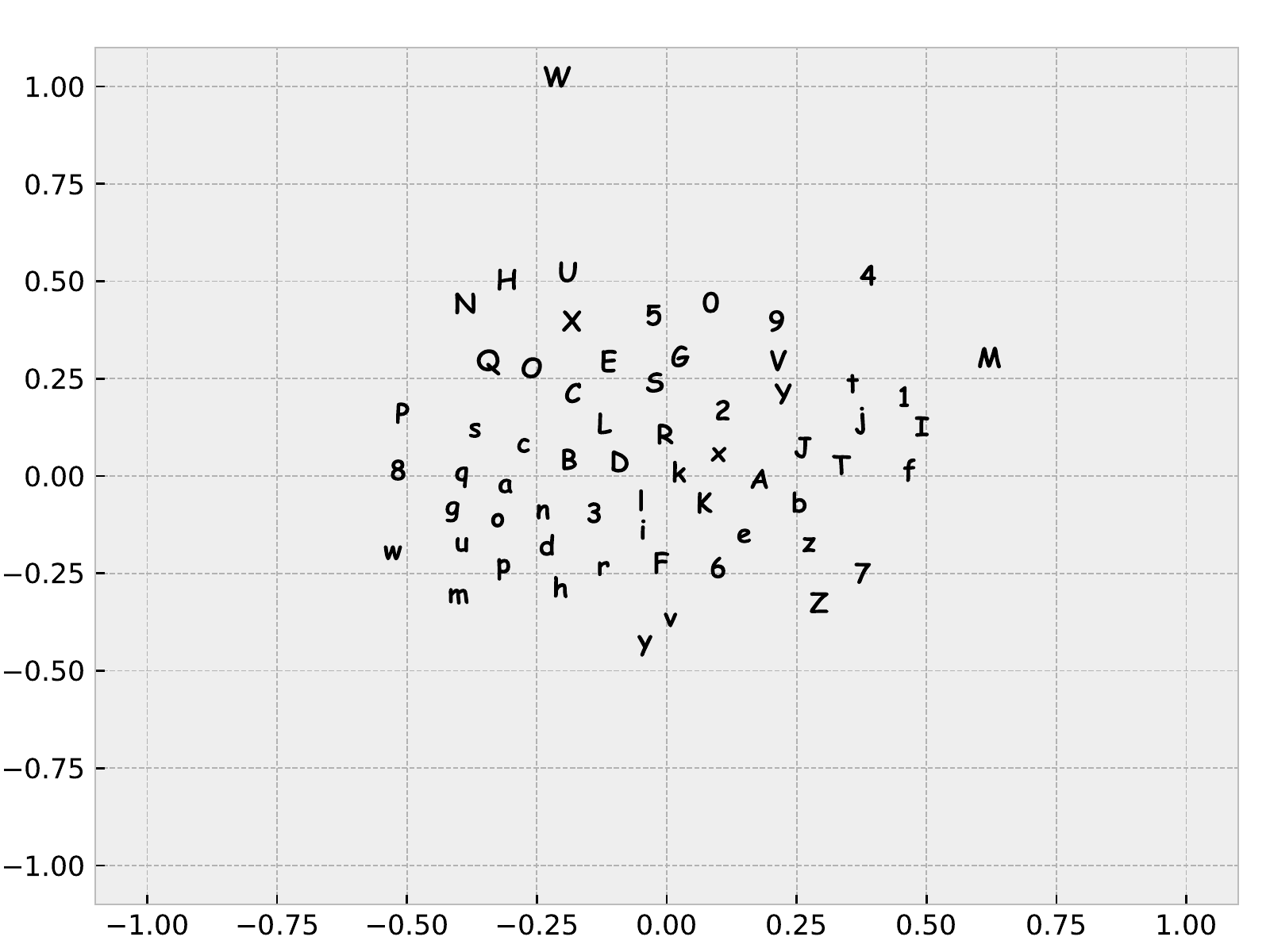}} 
    \subfigure[Times New Roman]{\includegraphics[width=0.24\textwidth]{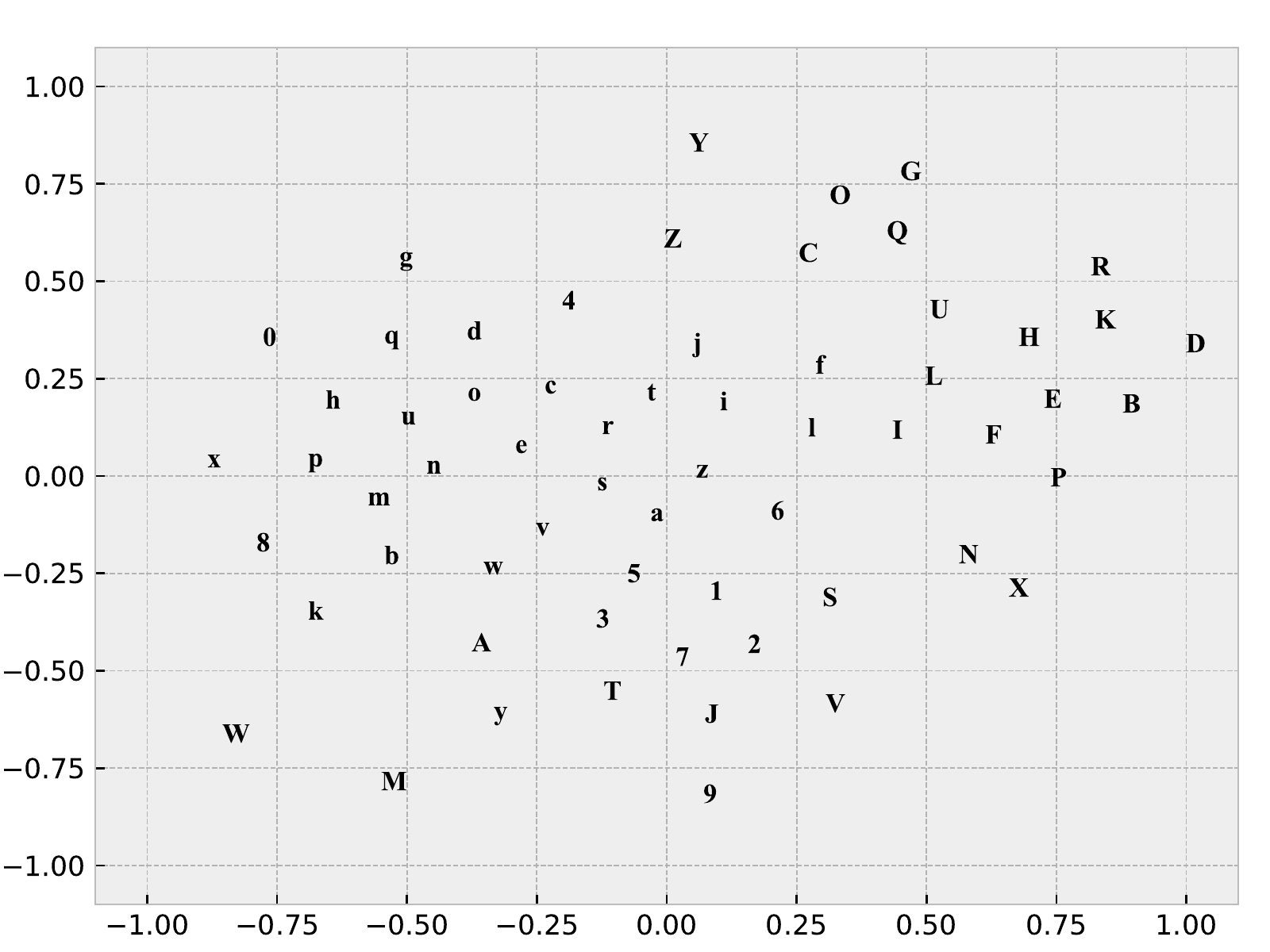}}
    \subfigure[Verdana]{\includegraphics[width=0.24\textwidth]{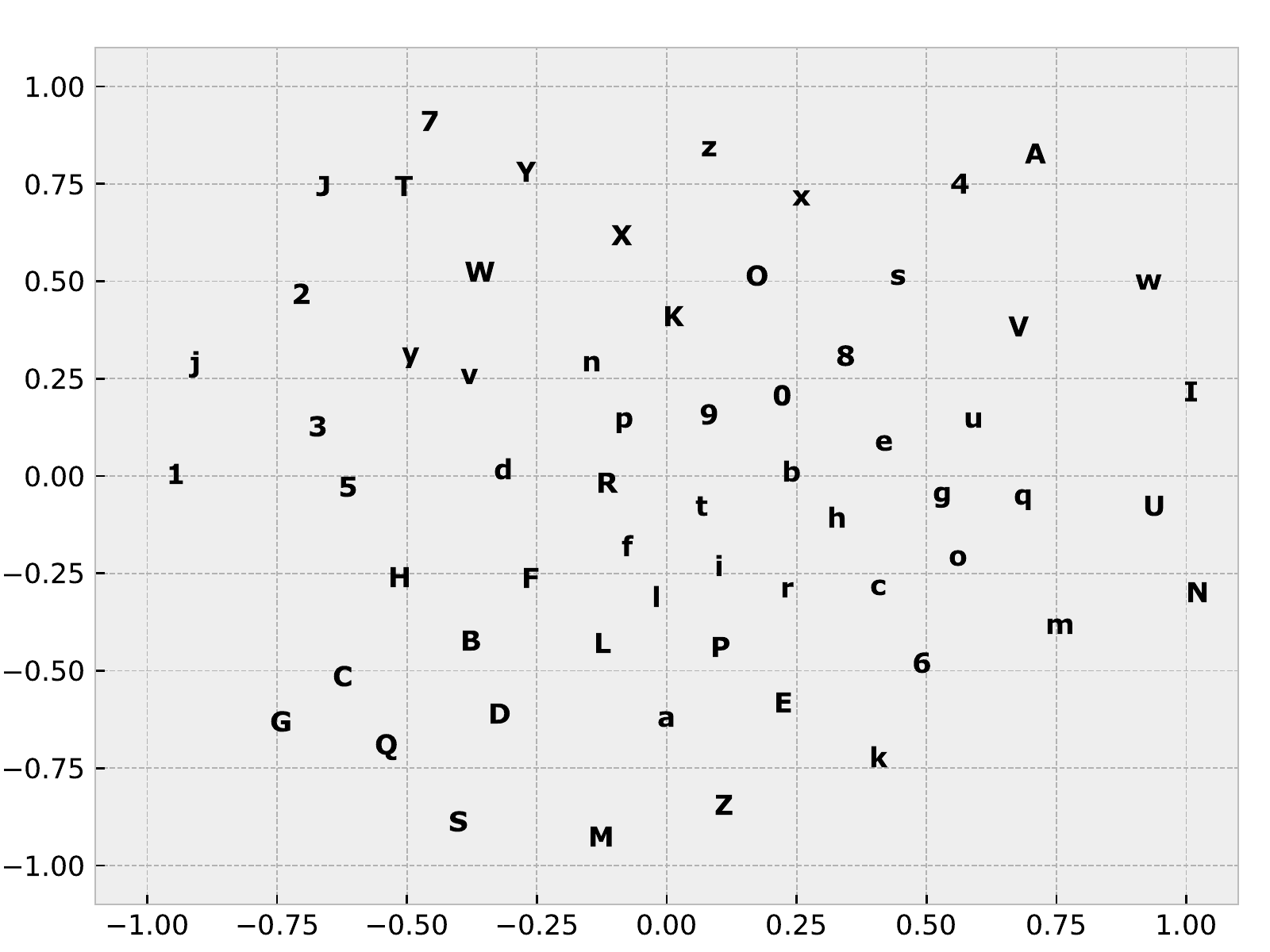}}
    \caption{Layout comparison of different \ve{} introduced by different fonts. The layouts appear different, but the proximity particulars are agreed across most fonts. (a) Arial (b) Comic Sans MS (c) Times New Roman (d) Verdana. Letters in the figures are printed in corresponding fonts for readers' convenience.}
    \label{fig:layout}
\end{figure*}

\paragraph{Fonts}
Additionally, 
to have a more thorough understanding of the choice of the fonts, we further plot the layout of the \ve{} with t-SNE in Figure~\ref{fig:layout}. 
Although the layouts appear distinct
given different fonts, 
the proximity particulars are roughly agreed. 
For example, the cluster of `C', `O', `Q', and `G', 
as well as the cluster of 
`e', `o', and `c' are agreed reasonably well 
across multiple fonts. 
Interestingly, we can see that
the embedding introduced by 
the font ``Comic Sans MS'' 
deviates the most from the layouts 
by other fonts, 
this deviation should be expected 
as letters printed with ``Comic Sans MS'' 
also look the most differently from other fonts. 

We also investigate whether our simple heuristic 
in determining the dimension $d$ will be affected 
significantly by the fonts. 
Our analysis suggest that 
there are barely any differences 
in terms of the function between $d$ and ratio of explained 
variance. 


\begin{table*}[]
\centering 
\small 
\begin{tabular}{ccccccccccc}
\hline
\multirow{3}{*}{$d$} & \multicolumn{2}{c}{\multirow{2}{*}{One-hot}} & \multicolumn{8}{c}{\emph{visual embedding}} \\
 & \multicolumn{2}{c}{} & \multicolumn{2}{c}{Arial} & \multicolumn{2}{c}{Comic Sans MS} & \multicolumn{2}{c}{Times New Roman} & \multicolumn{2}{c}{Verdana} \\ \cline{2-11} 
 & train & test & train & test & train & test & train & test & train & test \\ \hline
5 & \multirow{9}{*}{0.9784} & \multirow{9}{*}{0.8345} & 0.9453 & 0.9993 & 0.9224 & 0.9978 & 0.9433 & 0.9986 & 0.9269 & 0.9987 \\
10 &  &  & 0.9588 & 0.9996 & 0.9460 & 0.9990 & 0.9583 & 0.9996 & 0.9506 & 0.9994 \\
20 &  &  & 0.9389 & 0.9998 & 0.9427 & 0.9994 & 0.9549 & 0.9999 & 0.9606 & 0.9996 \\
30 &  &  & 0.9699 & 0.9994 & 0.9591 & 0.9995 & 0.9460 & 0.9999 & 0.9565 & 0.9990 \\
40 &  &  & 0.8842 & 0.9977 & 0.9707 & 0.9999 & 0.9450 & 0.9999 & 0.9548 & 0.9989 \\
50 &  &  & 0.968 & 0.9997 & 0.9478 & 0.9999 & 0.9552 & 0.9998 & 0.9473 & 0.9999 \\
60 &  &  & 0.9575 & 1.0000 & 0.9668 & 0.9999 & 0.9596 & 0.9999 & 0.9489 & 0.9999 \\
70 &  &  & 0.9453 & 1.0000 & 0.9454 & 0.9999 & 0.9639 & 0.9998 & 0.9562 & 0.9997 \\
80 &  &  & 0.9588 & 0.9998 & 0.9597 & 0.9998 & 0.9637 & 0.9999 & 0.9731 & 0.9999 \\ \hline
\end{tabular}
\caption{Synthetic experiments with simple robust text classification when \ve{} is tested with different font choices and dimensions. Classification accuracy is reported.}
\label{tab:synthetic}
\end{table*}

\subsection{Robust Machine Translation}

We proceed to examine the robustness of a NMT model trained over 
benchmark datasets against variation of inputs at test time. 
Following previous works \cite{belinkov2017synthetic, levy2019training}, we mainly used the IWSLT~2016 machine translation benchmark \cite{junczys2016university}, when the test data is polluted with various noises. 
We consider three language pairs: German-English (de-en), French-English (fr-en), and Czech-English (cs-en). 


\subsubsection{Competing Methods}
As the \ve{} is a generic 
method of encoding the characters of corpus, 
our method is expected to work 
with most of the other, if not all, character-level models. 
Therefore, we mainly test the baseline and the adversarial training methods, and compare them to the same methods when the 
input is \ve{}. 
To be specific, we study the following methods:

\begin{itemize}
    \item \base{}: The character-level transformer described in Section~\ref{sec:model}. 
    \item \bv{}: \base{} method when the input is \ve{}. 
    \item \adt{}: The input is the conventional one-hot embedding, but augmented with multiple synthetic noises, as done by \cite{belinkov2017synthetic,levy2019training}, this process is often discussed as adversarial training. In particular, we have the following five methods: 
    \begin{itemize}
        \item \adtd{}: when synthetic noises is to delete of a random character. 
        \item \adti{}: when synthetic noises is to insert a random character to a random position.
        \item \adtr{}: when synthetic noises is to replace a random character with another one.
        \item \adts{}: when synthetic noises is to swap two adjacent random characters.
        \item \adta{}: when synthetic noises is the combination of all above synthetic noises.
    \end{itemize}
    \item \adtv{}: the corresponding methods when input is \ve{}, correspondingly, we have \adtdv{}, \adtiv{}, \adtrv{}, \adtsv{}, and \adtav{}. 
\end{itemize}

Therefore, we tested 12 methods all together. Following \cite{levy2019training}, the methods concerning synthetic noises are only added to training data, the validation data remain untouched. Also, we follow the hyperparamter setup in \cite{levy2019training} to choose the probability $p=10\%$ in injecting synthetic noises. 


\subsubsection{Test Data with Noise}
Following precedents \cite{belinkov2017synthetic,levy2019training}. 
We consider two different situations 
for injecting noises into the test data:
the natural noise case
and the synthetic noise case. 

\paragraph{Natural Noise}
We evaluate the models where natural
noises are injected into the test data. 
We follow 
\cite{belinkov2017synthetic}
to add natural noises 
by leveraging the collection of 
frequently misspelled words 
and replacing the 
words in test sentences
with the misspelled ones. 
For each source language, we have: 
\begin{itemize}
    \item French: Wikipedia Correction and Paraphrase Corpus \cite{max2010mining}. 
    \item German: a combination of RWSE Wikipedia Revision Dataset \cite{zesch2012measuring} and The MERLIN corpus of language learners \cite{wisniewski2013merlin}
    \item Czech: manually annotated essays written by non-native speakers \cite{vsebesta2017czesl}. 
\end{itemize}

\paragraph{Synthetic Noise}
We also evaluate the models where different 
degrees of the synthetic noises are
injected into the test data. 
However, as the \adt{} family models are trained with synthetic noised injected to the text with probability $p=10\%$, 
we can intuitively expect these models will excel
at test time when the test sentences are perturbed with a similar degree of noises, 
even when these models are not in fact resilient towards variations in a broader scope. 
To avoid such inaccurate evaluations, 
we conduct a more comprehensive testing regime: 
for each model, we test it with all the different kinds of synthetic noises, and report the average score of these test performances:
therefore, even a model has seen certain noise pattern during the training phase, it may be fragile towards other noise patterns. 
Further, we also test beyond the setup where models are trained with the synthetic noises injected with probability $p$: 
we test the scenarios when the test sentences are perturbed with probability $2p$ and $3p$ in addition. 
For convenience of further discussion, we refer to this probability of which we inject synthetic noises as ``noise-level'' (NL). 

\subsubsection{Results}

\paragraph{Natural Text (Original Text and Natural Noise)}
We first evaluate the performances of these models in natural texts that can appear in the real world, 
including the original test texts 
and the misspelled texts generated by replacing the words with its frequently misspelled counterparts. 
Table~\ref{tab:main:all} shows the results, where natural texts are reported at the first two columns. 

Overall, we can see the best performances in these two categories are all obtained by methods with \ve{}, 
although the methods with the best performances in each category are different. 
In particular, we notice that \adtsv{} behaves reasonably well, 
it outperforms all the methods without \ve{} over the original text and the natural noises. 

If we compare the performances of the method and its counterpart with \ve{}, 
the improvement of \ve{} can be best represented in the \base{} method:
\bv{} almost maintains the performance 
of \base{} in original text
and shows an improvement
over the text with natural noises. 
Interestingly, we did not observe a clear change-of-performance pattern when \adt{} family models adopt \ve{}. 
In particular, some synthetic noises (deletion) hurt the convergence of \base{}, and \ve{} can recover (and improve) the performance, 
while some other synthetic noises (insertion and substitution) improves the performance over \base{}, but \ve{} degrades the performance. 
We run the experiments repeatedly, but find this performance pattern appears to be stable.


\begin{table*}[]
\centering 
\small 
\begin{tabular}{ccccccc}
\hline
\multirow{2}{*}{} & \multirow{2}{*}{original text} & \multirow{2}{*}{natural noise} & \multicolumn{3}{c}{synthetic noise} & \multirow{2}{*}{STD} \\
 &  &  & NL = $p$ & NL = $2p$ & NL = $3p$ &  \\ \hline
\base{} & 57.35 & 28.86 & 51.49 & 46.39 & 41.60 & 10.82 \\
\bv{} & 57.07 & 33.09 & \textbf{53.68} & 48.53 & 44.02 & 9.36 \\ \hline
\adtd{} & 48.6 & 38.32 & 43.31 & 40.64 & 38.05 & 4.36 \\
\adtdv{} & \textbf{63.55} & 51.01 & 49.47 & 45.88 & 42.90 & 7.92 \\ \hline
\adti{} & 60.84 & 50.13 & 51.70 & 39.73 & 37.97 & 9.38 \\
\adtiv{} & 49.19 & 39.61 & 41.48 & 49.62 & 47.85 & 4.66 \\ \hline
\adtr{} & 58.94 & 47.77 & 52.95 & 40.25 & 38.65 & 8.53 \\
\adtrv{} & 50.41 & 40.98 & 42.11 & \textbf{50.92} & \textbf{48.89} & 4.75 \\ \hline
\adts{} & 60.66 & 50.15 & 51.78 & 47.92 & 45.08 & 5.90 \\
\adtsv{} & 63.26 & \textbf{52.78} & 51.06 & 49.04 & 46.55 & 6.43 \\ \hline
\adta{} & 57.36 & 48.72 & 48.73 & 41.51 & 40.54 & 6.79 \\
\adtav{} & 49.65 & 42.72 & 42.33 & 47.44 & 46.10 & \textbf{3.12} \\ \hline
\end{tabular}
\caption{Test performances of French-English translation, where two models are reported together: the model uses one-hot embedding and \ve{}; performances are reported with text that can appear naturally (original text and natural noise) and text are perturbed with synthetic noises (with three different noise level (NL)); standard deviations (STD) of each row are also reported.}
\label{tab:main:all}
\end{table*}

\paragraph{Synthetic Noises}
Further, we inspect the models' robustness when the test sentences are perturbed by synthetic noises. 
Results are also showed in Table~\ref{tab:main:all}.

First of all, when we test the models with a more stringent setup, 
we get a message that counters the previous beliefs: 
our results indicate that 
training with synthetic noises alone does not necessarily increase the models' robustness against synthetic noises at test time.
In particular, although models trained with synthetic noises 
can outperform \base{} when the noise-level is $p$, 
the advantage of these models can be barely observed if we simply increase the noise-level. 

If we increase the noise-level, we can see a clear performance drop of all the models with one-hot embeddings, whether trained with synthetic noises or not. 
This behavior aligns with our conjecture:
synthetic noises can help improve the robustness of NMT models mostly because it offers an opportunity for the models to see the noise pattern during training, 
thus, when we shift the noise pattern (by increasing the noise-level), these models become fragile again. 

In contrast, models with \ve{} demonstrate an impressive level of resilience towards input variations as we increase the noise-level. 
Surprisingly, some models even see an increment of test score when the noise-level increases.  
For other models, as we increase the noise-level, the test score also goes down, but in a much slower pace than the models not boosted with \ve{}. 

The standard deviation also endorses the strength of \ve{}. 
With only one exception (\adtdv{}), 
all models with \ve{} report a smaller standard deviation, 
sometimes even smaller than half of the standard deviation of the counterparts. 

\paragraph{More Results from Appendix}
Due to the space limitation, we only report the detailed dissection of results on French-English translation. 
Results of other translations (German-English, Czech-English)
are reported in the Appendix~\ref{sec:app:all}. 
Briefly, all these results lead to the same conclusion:
\ve{} helps improve the robustness of NMT models towards input variations, 
especially when the noises during test phase are different from the ones used during train phase.

\section{Discussion}
\label{sec:diss}

\paragraph{Broader Usage of Visual Embedding:}
We believe our method can also be used beyond the MT 
task discussed, 
especially in the applications that constantly work with 
substandard inputs or compound words 
that are not even recorded in standard dictionaries. 

For example, 
we experiment with a collection of bioinformatics articles, 
over a task of sentiment analysis 
for drug reviews \cite{grasser2018aspect}. 
In biology research, 
there are usually new compound words created, 
therefore
it is likely that
the training corpus is inadequate to represent all the words a model will see during the test phase. 
Our results suggest that 
\ve{} leads to significantly better
results in comparison to the
previous deep learning driven methods. 
For example, a standard one-hot embedding model (CNN-LSTM architecture with highway network) 
can only achieve a classification accuracy around 0.35, while \ve{}
can boost the performance up to 0.72 with the same set of hyperparamters. 
As a reference, this recent manuscript reports an accuracy of 0.443 with a word-level neural network \cite{vskrlj2020tax2vec}. 

\paragraph{Broader Definition of Visual Embedding:}
Our paper describes the \ve{} in the context of encoding the shape of characters. 
However, one should be able to extend the concept to word level or even the sentence level (\textit{e.g.}, \cite{sun2018super,sun2019squared}). 

We have also attempted to extend the concept directly to the sentence or paragraph level: 
we directly print every sample into an image, and turn the NLP problem into a computer vision problem. 
However, we notice that this method is limited by the number of characters of a sample. 
For example, for a paragraph with 5000 characters, 
we roughly need an image with size $3000\times200$ to have all the characters printed clearly. 
The image will have 600k pixels and can easily surpass the 
capacity of most computer vision techniques. 
Converting the image into a lower-resolution one also fails as the blurry images will not be able to carry all the information the text has.  

\paragraph{Broader Scope of Related Work:}
We are not aware of related works that encoding the shape of input text in Germanic or Latin languages, especially in the context of learning robust NLP models. 
However, there is a large collection of related works that use the shape information of Sino-Tibetan languages, especially Chinese over various applications. 
We offer a brief overview of these works 
as they are connected to our paper in the sense of encoding the visual information of characters. 

For example, glyph-aware embedding has been explored to incorporate vision techniques to boost the language modelling and word segmentation task of Chinese \cite{DaiC17}. 
The visual encoding of Chinese also offers a convenience opportunity to investigate the radicals of the language, which as been taken advantage of for text classification and word segmentation \cite{ShiZYXL15}. Ideas in learning glyph embedding from vision representation has also been attempted for word analogy and word similarity in Chinese \cite{su2017learning,cao2018cw2vec}. 
Later, recent advances of this direction has enabled the usage of visual representation of characters, especially Chinese, in more high-level semantic applications \cite{LiuLLN17,MengWWLNYLHSL19,zhang2018which}.

\paragraph{Potential Limitations:}
Since our contribution, despite its simplicity, is a fundamental innovation that can be applied to nearly all the character-level models for nearly all the NLP tasks, 
we hope to discuss the potential limitations beyond specific application and model within the scope of this paper. 

First, although \ve{} is usually smaller than the one-hot embedding, and the model integrating the technique can usually finish one epoch faster, we notice that models using \ve{} needs more epoches to converge. 
Second, the usage of \ve{} may be limited in the applications when there is a undetermined effects of the shape of the words. 
For example, sometimes changing a few letters to upper case may even hinder the human (and \ve{}'s) recognition of it (\textit{e.g.}, banana vs. baNANA), but sometimes it may only hinder \ve{} but not human (\textit{e.g.}, banana vs. Banana). 
This misalignment may limit the usage of \ve{} for some dedicated applications. 
Fortunately, there should exist multiple heuristics that can account these issues to be explored, such as always mapping the first letter of a word to lower case.





\section{Conclusion}
\label{sec:con}
Motivated by the discrepancy between how humans 
and machine learning models process text data, 
we aim to improve the robustness of neural machine translation (NMT) models towards substandard inputs 
by aligning the data processing procedure between 
humans and the models. 
In particular, 
building around the belief that one of the reasons of the vulnerability of the models against substandard inputs is the discrepancy in processing data, 
we argue ``\emph{human read through eyes; so shall the models}.''

Following this argument, 
we introduced the \emph{visual embedding} (\ve{}), 
which encodes the shape of the characters 
as a replacement of the conventional one-hot embedding of characters. 
The \ve{} can be constructed very efficiently with only a few lines of codes, 
and 
it can be integrated into almost any existing character-level models that use one-hot embedding as input.

Further, in the context of machine translation over noised texts, 
we tested the performance of models with one-hot embedding and with \ve{}. 
We mainly compared to the methods that augment the training samples with noises (\textit{e.g.}, adversarial training), which is usually considered as one of the most effective methods for robust machine learning. 
With a more comprehensive evaluation, 
we demonstrated an impressive superiority of the \ve{}:
models with \ve{} (especially trained with synthetic noises) are resilient towards noised input 
even when the noises at the test time are introduced with a greater probability than that of the training phase, 
which is a scenario usually fails conventional methods. 
Overall, the empirical performance strongly endorsed the 
efficacy of \ve{}, especially in the context of robustness towards substandard inputs not seen during training.




\bibliography{ref}
\bibliographystyle{acl_natbib}

\appendix

\newpage 

\newpage 
\section*{Appendices}
\label{sec:appendix}







\section{Additional Results}
\label{sec:app:all}

\begin{table*}[]
\centering 
\small 
\begin{tabular}{ccccccc}
\hline
\multirow{2}{*}{} & \multirow{2}{*}{original text} & \multirow{2}{*}{nature noise} & \multicolumn{3}{c}{synthetic noise} & \multirow{2}{*}{STD} \\
 &  &  & NL = $p$ & NL = $2p$ & NL = $3p$ &  \\ \hline
\base{} & \textbf{55.45} & 26.76 & 31.05 & 29.78 & 28.46 & 11.93 \\
\bv{} & 46.47 & 29.57 & 34.72 & 33.15 & 31.38 & 6.66 \\ \hline
\adtd{} & 50.09 & 40.96 & 35.81 & 33.97 & 32.42 & 7.16 \\
\adtdv{} & 47.93 & \textbf{47.04} & 40.48 & 38.97 & 37.49 & 4.79 \\ \hline
\adti{} & 45.29 & 37.28 & 39.99 & 38.78 & 37.40 & 3.29 \\
\adtiv{} & 45.86 & 38.28 & 41.25 & 40.09 & 39.30 & 2.95 \\ \hline
\adtr{} & 52.73 & 37.44 & 39.98 & 38.60 & 37.50 & 6.50 \\
\adtrv{} & 45.2 & 43.03 & \textbf{45.50} & \textbf{43.89} & \textbf{42.38} & \textbf{1.35} \\ \hline
\adts{} & 49.58 & 38.4 & 40.91 & 39.20 & 37.83 & 4.83 \\
\adtsv{} & 46.69 & 40.21 & 42.06 & 40.34 & 38.80 & 3.06 \\ \hline
\adta{} & 50.78 & 38.22 & 36.60 & 35.99 & 35.58 & 6.42 \\
\adtav{} & 43.29 & 44.17 & 41.85 & 40.94 & 40.19 & 1.64 \\ \hline
\end{tabular}
\caption{Test performances of German-English translation, where two models are reported together: the model uses one-hot embedding and \ve{}; performances are reported with text that can appear naturally (original text and natural noise) and text are perturbed with synthetic noises (with three different noise level (NL)); standard deviations (STD) of each row are also reported.}
\label{tab:main:ge}
\end{table*}

\begin{table*}[]
\centering 
\small 
\begin{tabular}{ccccccc}
\hline
\multirow{2}{*}{} & \multirow{2}{*}{original text} & \multirow{2}{*}{nature noise} & \multicolumn{3}{c}{synthetic noise} & \multirow{2}{*}{STD} \\
 &  &  & NL = $p$ & NL = $2p$ & NL = $3p$ &  \\ \hline
\base{} & 46.31 & 30.78 & 44.02 & \textbf{41.91} & \textbf{39.83} & 5.98 \\
\bv{} & 38.87 & 32.91 & 38.37 & 37.31 & 36.02 & 2.38 \\ \hline
\adtd{} & 43.57 & 30.35 & 38.33 & 36.79 & 35.25 & 4.80 \\
\adtdv{} & \textbf{53.53} & 32.89 & \textbf{44.36} & 41.80 & 39.44 & 7.54 \\ \hline
\adti{} & 42.57 & 32.16 & 40.17 & 38.71 & 37.28 & 3.89 \\
\adtiv{} & 36.2 & 37.73 & 37.21 & 37.44 & 37.50 & \textbf{0.60} \\ \hline
\adtr{} & 43.61 & 32.29 & 39.98 & 38.56 & 37.14 & 4.14 \\
\adtrv{} & 36.48 & 33.88 & 37.54 & 36.85 & 36.09 & 1.39 \\ \hline
\adts{} & 46.69 & 31.64 & 38.95 & 37.41 & 35.84 & 5.52 \\
\adtsv{} & 49.69 & \textbf{34.14} & 38.23 & 38.32 & 38.28 & 5.85 \\ \hline
\adta{} & 36.77 & 28.9 & 22.56 & 20.82 & 19.54 & 7.15 \\
\adtav{} & 35.21 & 27.48 & 23.41 & 22.95 & 22.34 & 5.39 \\ \hline
\end{tabular}
\caption{Test performances of Czech-English translation, where two models are reported together: the model uses one-hot embedding and \ve{}; performances are reported with text that can appear naturally (original text and natural noise) and text are perturbed with synthetic noises (with three different noise level (NL)); standard deviations (STD) of each row are also reported.}
\label{tab:main:cz}
\end{table*}

We report the results of German-English translation and Czech-English translation here in Table~\ref{tab:main:ge} and Table~\ref{tab:main:cz} to support the discussion in the main manuscript. 
As expected, methods with \emph{visual embedding} surpass the counterparts with one-hot embedding 
over the noised sentences in most cases, although \emph{visual embedding} does not seem to help much in the original text. 
Interestingly, we noticed that \base{} model works pretty well in the Czech-English translation over especially when NL=$2p$ and NL=$3p$, we conjecture this performance is due to that other methods adopting adversarial training overfits the distribution of NL=$p$. 

Additionally, the column of STD shows that \emph{visual embedding} can significantly improve the robustness of the models towards variations of input. 

Overall, together with the results reported in the main manuscript, we can fairly conclude that \emph{visual embedding} is preferred over the one-hot embedding given the performances discussed. 


\end{document}